%% file: grad_norm.tex
\documentclass{article} % For LaTeX2e
\usepackage{iclr2016_conference,times}
\usepackage{hyperref}
\usepackage{url}
\usepackage{amsfonts}
\usepackage{bm}

\input{math_commands.tex}

\title{Efficient Per-Example Gradient Computations}

\author{Ian Goodfellow
\\
Google Inc., Mountain View, CA\\
\texttt{goodfellow@google.com} \\
}

\begin{document}

\maketitle

\begin{abstract}
This technical report describes an efficient technique for computing
the norm of the gradient of the loss function for a neural network
with respect to its parameters. This gradient norm can be computed
efficiently for every example.
\end{abstract}

\section{Introduction}

We often want to know the value of the $\normltwo$ norm of the gradient
of a loss function with respect to the model parameters for every example
in a minibatch. This is useful for techniques such as optimization based
on importance sampling~\citep{zhao2014stochastic}, where examples with large gradient norm
should be sampled more frequently. Unfortunately, most differentiation
functionality provided by most software frameworks does not support computing
gradients with respect to individual examples in a minibatch. This technical
report describes an efficient technique for obtaining all of the desired
gradient norms when training a standard neural network.

\section{Problem Definition}

Suppose that we have a neural network containing $n$ layers,
with each layer implementing a transformation
\[ \vz^{(i)} = \vh^{(i-1) \top} \mW^{(i)} \]
\[ \vh^{(i)} = \phi^{(i)} ( \vz^{(i)} ).\]

Here $\vh^{(i)}$ are the activations of the hidden layers of
the neural network. To simplify notation, we refer to the input
of the network as $\vh^{(0)}$. The activation function $\phi$
may be any differentiable function that does not contain any
model parameters (for example, it need not be element-wise).
The matrix $\mW^{(i)}$ is the weight matrix for layer $i$. To
simplify the presentation, we treat the biases for each layer
as being an extra column of $\mW$, with the $\phi$ function from the layer
below providing a constant input of $1$ to this column.

Suppose further that we have a loss function
\[ L(\vz^{(1)}, \dots, \vz^{(n)}, \vh^{(0)}, \vy) \]
where $\vy$ is a vector of targets provided from the training set.
Note that $L$ is this a function of the targets and the activations
of the neural network, but is not permitted to access the model parameters
themselves. The model parameters are accessed only via their influence
on $\vz$.

Given a minibatch of $m$ examples, define $L^{(j)}$ to be the loss
when using example $j$ to provide $\vh^{0}$ and $\vy$, and define
the total cost $C$ to be the sum of all $m$ of these values.

Our goal is to compute $\vs^{(i)}$ for $i=1, \dots, n$, where $\vs^{(i)}$ is
a vector of the sum of squared derivatives, defined by
\[\evs^{(i)}_j= \sum_{k,l} (\frac{\partial}{\partial \emW^{(i)}_{k,l}} L^{(j)})^2\].

The $\normltwo$ norm of the parameter gradient for example $j$
is then given by $\sqrt{\sum_i \evs^{(i)}_j}$. Other norms, for example,
the norm of the gradient for an individual weight matrix, can also be
computed easily from the $\vs$ vectors.

\section{Naive approach}

Back-propagation
(the term here refers not so much to the original backpropagation
algorithm as to any of the modern general symbolic or automatic
differentiation algorithms)
allows the computation
of $\frac{\partial}{\partial \emW_{k,l}} C$.
The naive approach to computing $\vs$ is to run back-propagation $m$
times with a minibatch size of 1. This provides the gradient for each
example. Each of these per-example gradients may then be summed out
explicitly.

\section{Proposed method}

Let $\mH^{(i)}$ be a matrix representing a minibatch of values of
$\vh^{(i)}$, with row $j$ of $\mH^{(i)}$ containing $\vh^{(i)}$ for
example $j$. Let $\mZ^{(i)}$ likewise contain a minibatch of $\vz^{(i)}$
values.

Standard backpropagation values allow us to compute the gradient of $C$
with respect to all of the $\mZ$ matrices in a single pass. Let
$\bar{\mZ} = \nabla_\mZ C$.

We then have
\[\vs^{(i)}_j = \left(\sum_k (\bar{\emZ}^{(i)}_{j,k})^2 \right)\left(\sum_k (\emH^{(i-1)}_{j,k})^2\right).\]

\section{Comparison}

Suppose that each layer has dimension $p$. Then the asymptotic number of
operations required by
back-propagation for computing the parameter gradients is $O(mnp^2)$.
The naive method for computing $\vs$ also uses $O(mnp^2)$ operations.
The naive method does not re-use any of the computations used by back-propagation,
so it roughly doubles the number of operations.
The new proposed method re-uses the computations from back-propagation.
It adds only $O(mnp)$ new operations. For large $p$, the extra cost of using
the proposed method is thus negligible.
In practice, the difference between the methods is much greater than this
asymptotic analysis suggests. The naive method of running back-propagation $m$
times with a minibatch of 1 performs very poorly because back-propagation is
most efficient when efficient matrix operation implementations can exploit
the parallelism of minibatch operations.

\section{Extensions}

Computing $\vs$ allows other per-example gradient operations. For example,
after determining the norm of the gradient for each example, we can modify
the original $\bar{\mZ}$ values, for example, by rescaling each row to
satisfy a constraint on the norm of the parameter gradient. After obtaining
the new modified $\mZ'$ we can re-run the final step of backpropagation, to
obtain
\[
\bar{\mW}^{(i)'} = \mX^{(i)\top}\bar{\mZ}^{(i)'}.
\]

\bibliography{ml,aigaion}
\bibliographystyle{iclr2016_conference}

\end{document}

%% file: math_commands.tex
\def\1{\bm{1}}

\def\vh{{\bm{h}}}

\def\vs{{\bm{s}}}

\def\vy{{\bm{y}}}
\def\vz{{\bm{z}}}

\def\evs{{s}}

\def\mH{{\bm{H}}}

\def\mW{{\bm{W}}}
\def\mX{{\bm{X}}}

\def\mZ{{\bm{Z}}}

\DeclareMathAlphabet{\mathsfit}{\encodingdefault}{\sfdefault}{m}{sl}
\SetMathAlphabet{\mathsfit}{bold}{\encodingdefault}{\sfdefault}{bx}{n}

\def\emH{{H}}

\def\emW{{W}}

\def\emZ{{Z}}

\newcommand{\normltwo}{L^2}